\documentclass[conference]{IEEEtran}
\IEEEoverridecommandlockouts
\usepackage{cite}
\usepackage{amsmath,amssymb,amsfonts}
\usepackage{algorithmic}
\usepackage{graphicx}
\usepackage{textcomp}
\usepackage{csvsimple}
\usepackage{booktabs}
\usepackage{xcolor}
\usepackage[T1]{fontenc}

\newcommand{\reftable}[1]{Table~\ref{#1}}

\newcommand{\argmax}{\operatornamewithlimits{argmax}} 

\def\BibTeX{{\rm B\kern-.05em{\sc i\kern-.025em b}\kern-.08em
    T\kern-.1667em\lower.7ex\hbox{E}\kern-.125emX}}
\begin{document}

\title{General Game Heuristic Prediction Based on Ludeme Descriptions\\
\thanks{This research is funded by the European Research Council as part of the Digital Ludeme Project (ERC Consolidator Grant \#771292) led by Cameron Browne at Maastricht University's Department of Data Science and Knowledge Engineering.}
}

\author{\IEEEauthorblockN{Matthew Stephenson, Dennis J. N. J. Soemers, {\'E}ric Piette, Cameron Browne}
\IEEEauthorblockA{\textit{Department of Data Science and Knowledge Engineering} \\
\textit{Maastricht University}\\
Maastricht, the Netherlands \\
\{matthew.stephenson,dennis.soemers,eric.piette,cameron.browne\}@maastrichtuniversity.nl}
}

\maketitle

\begin{abstract}
This paper investigates the performance of different general-game-playing heuristics for games in the Ludii general game system. Based on these results, we train several regression learning models to predict the performance of these heuristics based on each game's description file.
We also provide a condensed analysis of the games available in Ludii, and the different ludemes that define them.
\end{abstract}

\begin{IEEEkeywords}
Ludii, Heuristics, Ludemes, General Game Playing, Data Mining, Supervised Learning
\end{IEEEkeywords}

\section{Introduction}

Ludii is a general game system focused on traditional board games \cite{Piette2020Ludii}. It is being developed as part of the Digital Ludeme Project \cite{ludii1} but also provides opportunities for stand-alone AI research in the areas of agent development, content generation, game design, player analysis and education. 
Games within Ludii are described using ludemes, which are unique tokens or keywords that typically correspond to high-level game concepts. 
In this paper, we investigate if the performance of different game-playing heuristics can be estimated based on this ludemic game description format. If successful, this will allow us to predict the best heuristic for any new game, without the need for lengthy performance comparisons.

Many game-playing agents rely on heuristics to determine the best action to perform at a given time. In Chess, for example, one heuristic may encourage an agent to gain a material advantage, while another a positional advantage.
Techniques that aim to predict the best choice from a collection of heuristics are commonly called hyper-heuristic approaches \cite{hyperSurvey}.
Such approaches have been proposed for different games and research frameworks, including the general video game AI (GVGAI) framework \cite{Bontrager2016MatchingGA,7860398}, the general game playing (GGP) framework \cite{6571225}, Starcraft \cite{6633643}, Angry Birds \cite{AIIDE1715828}, Pac-Man \cite{8490401}, Jawbreaker \cite{SalcedoSanz2013AnEH}, FreeCell \cite{6249736}, and theoretical games \cite{7017583}.

The remainder of this paper is structured as follows. Section II describes the games and their ludemic format within Ludii. Section III describes the game-playing heuristics. Section IV describes the heuristic prediction methodology, experiments and results. Section V provides a short discussion of these results and possibilities for future work. 

\section{Ludii Games}

The Ludii general game system currently includes over 750 fully playable games.\footnote{All values presented in this paper were obtained using Ludii v1.1.17.} The types of games within Ludii includes, but is not limited to:
\begin{itemize}
  \item Deterministic / Stochastic Games
  \item Complete / Hidden Information Games
  \item Single-Player / Multi-Player Games
  \item Alternating / Simultaneous Move Games
\end{itemize}
For this paper, we will focus only on games that require two or more players, have an alternating move format, and are fully observable (no hidden information). This subset includes a total of 695 valid games.

\subsection{Ludemes}

Games within Ludii are described in terms of distinct ludemes.
The number of ludemes used within each description can range from a few dozen for simple games (e.g. Tic-Tac-Toe) to several thousand for larger games (e.g. Taikyoku Shogi). 
Each ludeme represents a fundamental aspect of play, allowing for high-level structured descriptions. For example, the ludemic description for Tic-Tac-Toe is as follows:

\footnotesize
\begin{verbatim}
  (game "Tic-Tac-Toe"  
      (players 2)  
      (equipment { 
          (board (square 3)) 
          (piece "Disc" P1) 
          (piece "Cross" P2) 
      })  
      (rules 
          (play (move Add (to (sites Empty))))
          (end (if (is Line 3) (result Mover Win)))
      )
  )
\end{verbatim}
\normalsize

Keywords such as Square, Piece, Line, etc. represent ludemes. Values defined by Strings, Numbers and Booleans (e.g. "Disc", 3, and True) are not considered ludemes.

The set of ludemes within the Ludii Language is continuously expanding but currently contains 547 ludemes. 
Some are used commonly, e.g. \emph{Board} and \emph{Players} are present in almost every game, while others are far less frequent, e.g. \emph{Liberties} and \emph{Enclose} are mostly found in Go-like games. 
By recording which ludemes are present within each game's description, we can produce a complete ludeme dataset.

\subsection{Ludii Game Clusters}
\begin{figure}
\centerline{\includegraphics[width=0.95\linewidth]{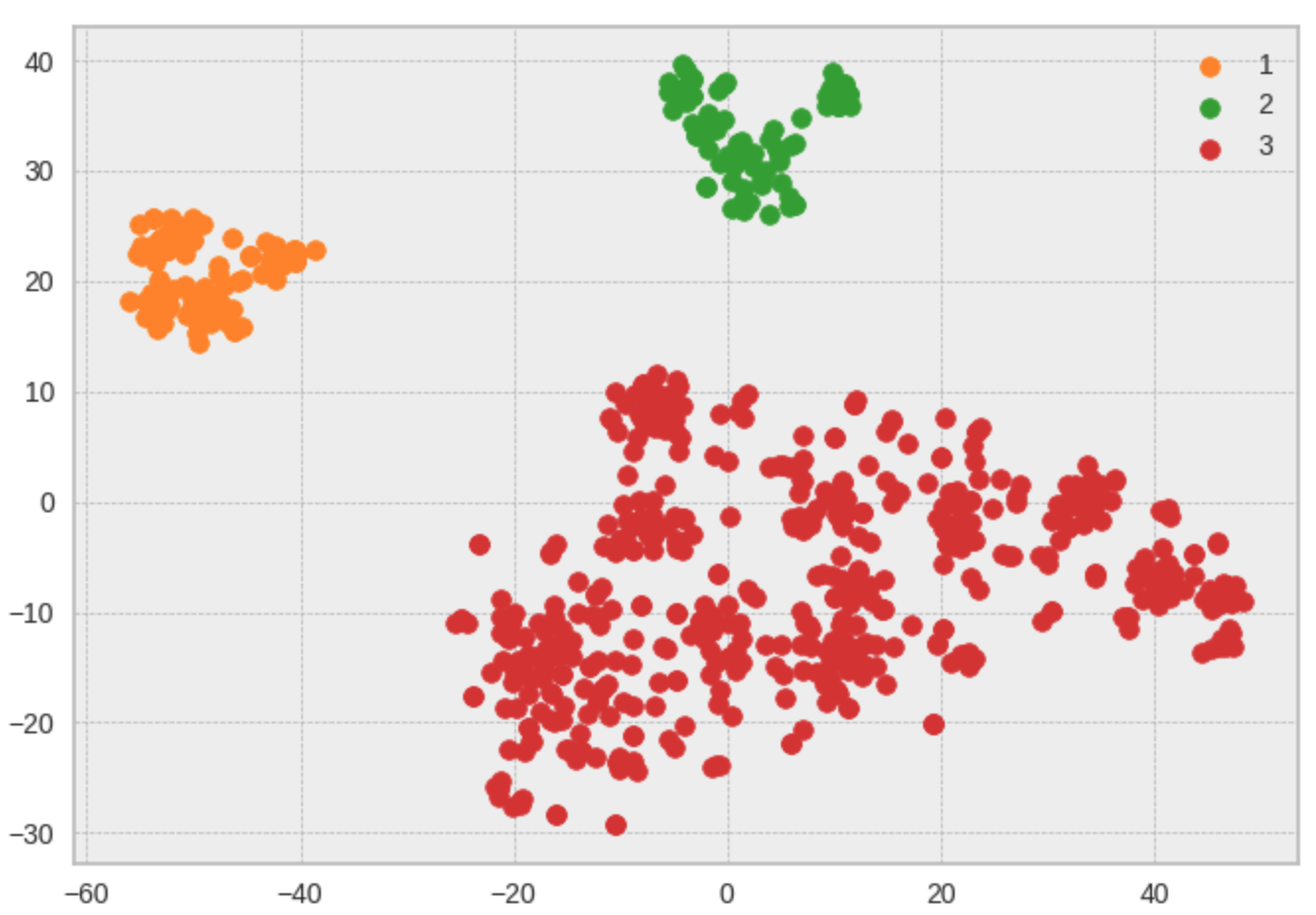}}
\caption{Example game clusters based on our ludeme dataset. Points were reduced to two dimensions using t-SNE.}
\label{gameClusters}
\end{figure}

To get a better idea of the overall distribution of the games available in Ludii, we applied t-distributed stochastic neighbor embedding (t-SNE) \cite{JMLR:v9:vandermaaten08a} to reduce our ludeme dataset to two-dimensions, see Figure \ref{gameClusters}. 
From this, we can see 3 distinct clusters of games. The orange cluster contains 88 games, the green cluster contains 85 games, and the red cluster contains 522 games.
Using a decision tree classifier, we identified which ludemes could be used to define this clustering.

The orange cluster contains games which have the \emph{Sow} ludeme, a ludeme which is specific to the ``Mancala" family of games. These games have a unique play-style, as they do not rely on traditional ideas of piece movement, placement and capturing in the same way that many other games do. Instead, players take turns ``sowing" seeds along a ring of holes, often with many seeds located on a single space.

The green Cluster contains games which have both the \emph{Track} and \emph{Dice} ludemes. This is a common combination for the ``Race" family of games, which includes popular examples such as Snakes \& Ladders and Backgammon. The \emph{Dice} ludeme means that dice are used as part of the game's rules, and would indicate that there is some aspect of luck to the game. The \emph{Track} ludeme indicates that there are fixed paths on the board which pieces must follow, often limiting the number of possible moves that can be made. Both of these ludemes together would suggest that the game has a limited degree of interaction, with the outcome of a player's turn often being determined more by chance than skill.

The red cluster contains all other games which did not fit into either of the previous clusters, including games that have a track but no dice (e.g. Ja-Jeon-Geo-Gonu) and games with dice but no track (e.g. Dice Chess). It is likely that there other additional, albeit less clearly separated, clusters within this large set of games, but this was not investigated as part of this preliminary analysis.

\begin{table*}[t]
\caption{Heuristics implemented in Ludii. The second-to-last and last columns show the win-rates of the heuristics averaged over all games (and number of games in which a heuristic is the sole top performer), for cases where the heuristics are assigned a positive or negative weight, respectively.}
\label{Table:HeuristicsList}
\begin{center}
\small
\begin{tabular}{@{}llrr@{}}
\toprule
& & \multicolumn{2}{c}{Avg. Win $\%$ (\# top performances)} \\
     \cmidrule(lr){3-4}
Heuristic              & Description & Positive    & Negative   \\
\midrule
Material               & Sum of owned pieces. & 62.09\% (181)       & 39.88\% (12) \\
Mobility         & Number of legal moves. & 57.75\% \space(65)  & 43.58\% (13)  \\
Influence              & Number of legal moves with distinct destination positions. & 57.59\% \space(53)  & 44.70\% \space(6)  \\
CornerProximity        & Sum of owned pieces, weighted by proximity to nearest corner. & 57.27\% \space(44)  & 40.92\% \space(9) \\
SidesProximity         & Sum of owned pieces, weighted by proximity to nearest side. & 56.36\% \space(31)  & 43.86\% (14) \\
LineCompletion         & Measure of potential to complete line(s) of owned pieces. & 54.86\% \space(73)  & 44.28\% \space(3)  \\
CentreProximity        & Sum of owned pieces, weighted by proximity to centre. & 54.53\% \space(37)  & 44.59\% (13) \\
RegionProximity        & Sum of owned pieces, weighted by proximity to a predefined region. & 53.07\% \space(21)  & 47.60\% \space(7)  \\
OwnRegionsCount        & Sum of (piece) counts in owned regions. & 50.35\% \space(19)  & 47.66\% (10) \\
PlayerRegionsProximity & Sum of owned pieces, weighted by proximity to owned region(s). & 50.01\% \space\space(2)   & 49.50\% \space(2)   \\
PlayerSiteMapCount     & Sum of (piece) counts in sites mapped to by player ID. & 49.99\% \space(18)  & 47.86\% \space(2)  \\
Score                  & Score variable of game state corresponding to player. & 49.83\% \space(17)  & 47.39\% \space(0)  \\
ComponentValues        & Sum of values of sites occupied by owned pieces. & 48.68\% \space\space(0)   & 48.39\% \space(0)  \\
Null                   & Always returns a value of $0$. & 48.52\% \space\space(0)   & -          \\
\bottomrule
\end{tabular}
\label{tab1}
\end{center}
\end{table*}

\section{Ludii AI Heuristics}
Ludii contains several heuristics that may be used in heuristic state evaluation functions for game-playing algorithms that require them, such as $\alpha\beta$-search \cite{Knuth1975AlphaBeta}. The selection of heuristics implemented in Ludii is primarily based on the advisors that were found to be generally applicable and useful (not just for a specific game) in the earlier Ludi general game system \cite{Browne2009}. The first two columns of \reftable{Table:HeuristicsList} list names and short descriptions of each of them. For every heuristic, we include a positive and a negative variant, with positive and negative weights, respectively. For example, Material with a positive weight prefers game states where a player owns more pieces (in comparison to the opponents), whereas Material with a negative weight prefers game states where the player owns fewer pieces.

\subsection{Heuristic Performance}
For each of the 695 games, we aim to get an estimate of the ``usefulness'' of each heuristic for that game. This is done by measuring the average win-rates of $\alpha\beta$-search agents using each heuristic against combinations of all other heuristics. To make this study feasible across many games, we restricted the agents to search depths of $2$. Furthermore, we exclude heuristics from games that they are not applicable in, and instead estimate what their performance would have been by the performance of the ``Null'' heuristic, which always returns a constant value of $0$. For example, the ``Score'' heuristic is inapplicable in games that do not use scores, ``RegionProximity'' is inapplicable in games that do not have any defined regions, etc.

More concretely, given an $n$-player game with a pool of $k$ applicable heuristics, for every heuristic $h$, we generate all combinations of $n - 1$ heuristics out of the other $k - 1$ as opponents. If there are more than $10$ such combinations (in games with many players and many applicable heuristics), we randomly sample only $10$ combinations to retain. We play a minimum of $10$ evaluation games for every such heuristic $h$ against every combination of opponents, with a minimum of $100$ evaluation games per heuristic $h$ (plus more games when $h$ itself is in the pool of candidate opponents for other heuristics).

The average win-rate of each heuristic across all 695 games is shown in \reftable{Table:HeuristicsList}, for both the positive and negative variants. The number shown in brackets is the number of times this heuristic had the exclusive highest win-rate on a game (i.e. the number of games where it outperformed all other heuristics). Some heuristics (e.g. Material, Mobility, Influence) with positive weights have relatively high average win-rates, which suggests these heuristics may be good candidates for ``default'' heuristics in new games with zero domain knowledge. Almost all heuristics -- including several with average win-rates around $50\%$ -- are still the top performers in some games, which suggests they are valuable to retain in a portfolio of general game heuristics. When using negative weights, all heuristics appear to have a poor level of performance on average, but most of them are still top performers in some games.

\section{Heuristic Performance Prediction}

To determine if a game's ludemes can be used to predict heuristic performance, we experimented with several regression learning algorithms. The names of these algorithms can be seen in Table \ref{tab2}. We also tested a naive approach, which always predicts the mean of all label values.
For each regression algorithm, a different model was independently trained for each heuristic. Each of these algorithm-heuristic models was trained across all 695 games, using the heuristic's win-rate as the labels and the ludemes as the input features.

All results for this section are shown in Table \ref{tab2}. All model training was performed in Python (v3.8.5) using Scikit-learn (v0.24.1) \cite{scikit-learn}, with all regression learning algorithms trained using default hyperparameter values. All of the data and code used in these experiments is open-source and available online.\footnote{https://github.com/Ludeme/HeuristicPrediction/tree/CoG\_2021} 

\subsection{Accuracy}

To determine the accuracy of each trained model, we performed Leave-One-Out Cross-Validation and recorded the Mean Average Error (MAE) of each model. Due to space limitations, we could not show the MAE of each separate algorithm-heuristic model, but instead show the average MAE of each regression algorithm across all heuristics. RandomForestRegressor performs best overall, with an MAE of $7.48$.

\subsection{Expected Win-Rate}

To get a better idea of how these models would compare if used by a portfolio agent, we also calculated the average expected win-rate of each regression algorithm. The expected win-rate of a regression algorithm on a given game, is equal to the true win-rate of its most preferred heuristic (i.e. the heuristic with the highest predicted win-rate) for that game. The average expected win-rate is computed by averaging over all games. RandomForestRegressor once again performs best, with an average expected win-rate of $71.03\%$.

\subsection{Regret}

For a single game, the regret of a regression algorithm is defined as the expected reduction in win-rate as a result of picking its most preferred heuristic over whichever heuristic truly performs best. An increase in regret is inversely proportional to that of expected win-rate, but shows how close each regression algorithm is to perfect heuristic prediction.
More formally, the regret $\mathcal{R}(h)$ for a regression algorithm that picks a heuristic $h$ with a win-rate $w(h)$, in a game in which the best heuristic $h^* = \argmax_{h'} w(h')$ has a win-rate $w(h^*) \geq h$, is given by $\mathcal{R}(h) = w(h^*) - w(h)$. The average regret is computed by averaging over all games. RandomForestRegressor has an average regret of $7.68$.

\section{Discussion}

Based on the results shown in Table \ref{tab2}, the best performing regression algorithm (RandomForestRegressor) had less than half as much regret as the naive approach, representing a significant performance improvement. While not as large, the reduction in MAE was also substantial. This result confirms that the ludemes within a game's description can be used to predict the performance of our heuristics, and that these predicted win-rates can be used effectively to select the best performing heuristic on new games.

There is also a very high correlation between the MAE and Regret results, with a lower MAE almost always correlating with a lower Regret. However, the improvement against the naive approach was much greater for regret than for accuracy (54\% improvement versus 26\% improvement). This would suggest that our prediction models are generally more accurate for games where the heuristics perform very differently, compared to cases where the performance is similar. In other words, if a game has low variance in terms of heuristic win-rate, then having a large MAE is typically going to have much less of an impact on our regret than if the game had high variance in heuristic win-rate.

\begin{table}[t]
\caption{Heuristic Performance prediction results. MAE values for each algorithm are averaged across all heuristics.}
\begin{center}
\small
\begin{tabular}{@{}lrrr@{}}
\toprule
Regression Algorithm              & MAE (stdev) & Win-Rate & Regret\\
\midrule
RandomForestRegressor           & 7.48 (1.28) & 71.03\% & 7.68 \\
GradientBoostingRegressor       & 7.87 (1.54) & 70.83\% & 7.88 \\
KNeighborsRegressor             & 8.05 (1.39) & 70.46\% & 8.25 \\
BayesianRidge                   & 8.29 (1.50) & 70.26\% & 8.45 \\
LinearSVR                       & 8.56 (1.31) & 69.97\% & 8.74 \\
MLPRegressor                    & 8.90 (1.40) & 70.15\% & 8.56 \\
ElasticNet                      & 9.13 (1.83) & 67.92\% & 10.79\\
Lasso                           & 9.26 (1.81) & 67.63\% & 11.08\\
Ridge                           & 9.50 (1.76) & 68.86\% & 9.85\\
DecisionTreeRegressor           & 9.70 (1.61) & 67.95\% & 10.76\\
Naive                           & 10.07 (2.58) & 62.09\% & 16.62\\
\bottomrule
\end{tabular}
\label{tab2}
\end{center}
\end{table}

\subsection{Future Work}

In this section we briefly describe several preliminary ideas for future work, including both extensions to and applications of the research proposed in this paper.

\subsubsection{Game Concepts}

Rather than using only the ludemes within our game descriptions, including additional game features or concepts may provide better results. Several higher level features have been previously proposed for general video games \cite{Bontrager2016MatchingGA,7860398}, but these are limited to a few dozen examples and are typically not applicable to board games. 
We have already begun work on creating a corpus of game concepts that are detectable within Ludii games \cite{arxivConceptsPaper}. Combining these concepts with our existing ludeme dataset may help to improve our heuristic prediction accuracy, and could allow us to extend our results to other general game systems.

\subsubsection{Combining Heuristics}

While the method described here identifies a single best predicted heuristic per game, better agent performance can typically achieved by using a weighted combination of several heuristics. 
For example, Material and Mobility tend to combine well for many games to encourage game states in which a player has a material advantage over their opponent(s) and more movement options. 
Unfortunately, the large number of potential combinations of heuristics, and the fine tuning of their weights, makes the identification of optimal weight vectors a difficult task for hundreds of games.

\subsubsection{Portfolio Agent}

One of the primary applications of hyper-heuristic approaches is the development of a portfolio agent. These agents have access to a wide range of different heuristics or agents, and select the best one for a given game. For our case, the general performance of a basic portfolio agent which selects a heuristic based on our RandomForestRegressor can be estimated from our expected win-rate. However, there are several other aspects to portfolio agent development which can also be considered, such as selecting an alternative heuristic should the first choice perform poorly.

\subsubsection{Strategy Recommendation}

Each of our heuristics is simple enough that we can use their predicted win-rates to provide simple strategies or tutorials for human players to follow. For example, the game of Go has high predicted win-rates for the Material (positive) and Score (positive) heuristics. Certain models, such as decision trees, may also be able to provide which ludemes are responsible for these predictions. This information could then be combined into a single complete message that provides some strategic insight, such as ``This game contains the Remove and AddScore ludemes, you should focus on gaining a Material and Score advantage".

\bibliographystyle{IEEEtran}
\bibliography{references}

\end{document}